\documentclass[10pt,twocolumn,letterpaper]{article}

\usepackage[pagenumbers]{cvpr}
\usepackage{times}
\usepackage{epsfig}
\usepackage{graphicx}
\usepackage{amsmath}
\usepackage{amssymb}
\usepackage{url}
\usepackage{booktabs}
\usepackage{amsfonts}
\usepackage{nicefrac}
\usepackage{microtype}
\usepackage{comment}
\usepackage{multirow}
\usepackage{wrapfig}
\usepackage{float}
\usepackage{multicol}
\usepackage{dblfloatfix}
\usepackage[font=small,labelfont=bf]{caption}
\usepackage[page]{appendix}

\usepackage[dvipsnames]{xcolor}

\definecolor{cvprblue}{rgb}{0.21,0.49,0.74}
\usepackage[pagebackref,breaklinks,colorlinks,citecolor=cvprblue]{hyperref}

\begin{document}

\title{FutureHuman3D: Forecasting Complex Long-Term \\3D Human Behavior from Video Observations}

\author{%
  Christian Diller \\
  Technical University of Munich\\
  {\tt\small christian.diller@tum.de} \\
   \and
   Thomas Funkhouser \\
   Google \\
   {\tt\small tfunkhouser@google.com} \\
   \and
   Angela Dai \\
   Technical University of Munich \\
   {\tt\small angela.dai@tum.de} \\
}

\twocolumn[{%
    \vspace{-4.5em}
	\renewcommand\twocolumn[1][]{#1}%
	\maketitle
	\begin{center}
    \includegraphics[width=0.9\linewidth]{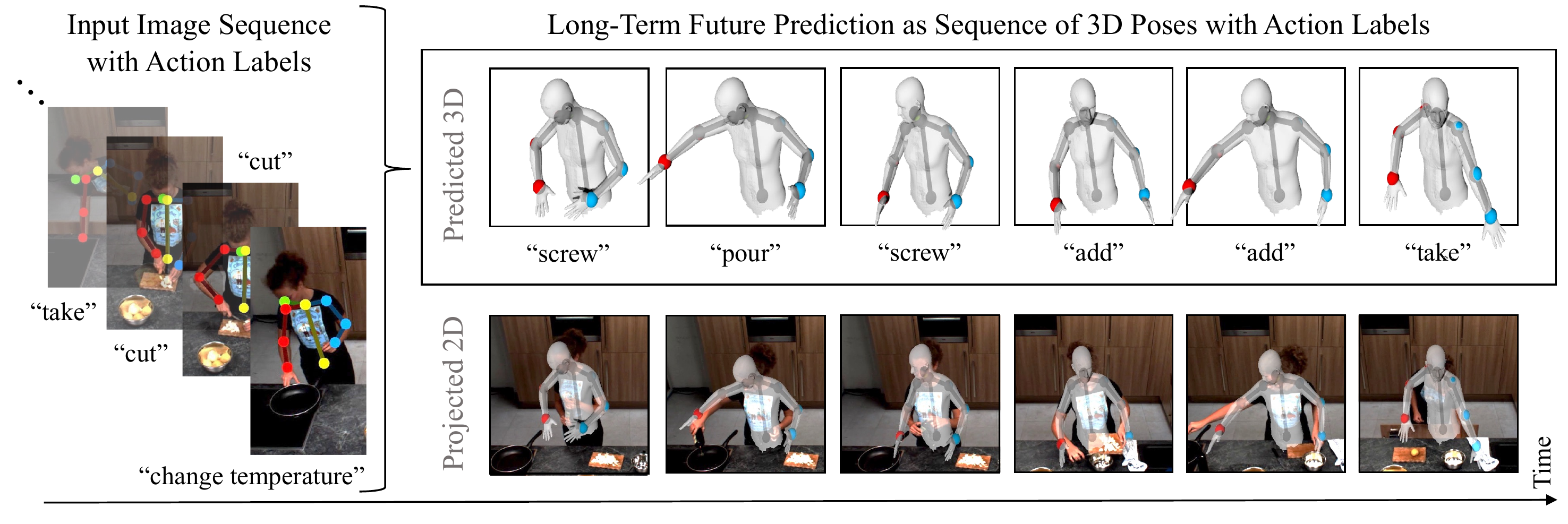}
    \vspace{-1em}
    \captionof{figure}{
    We propose a novel generative approach to model long-term future human behavior by jointly forecasting a sequence of coarse action labels and their concrete realizations as 3D body poses. For broad applicability, our autoregressive method only requires weak supervision and past observations in the form of 2D RGB video data, together with a database of uncorrelated 3D human poses.
    }
    \label{fig:teaser}
	\end{center}
}]

\begin{abstract}
\label{sec:abstract}
\vspace{-1em}

We present a generative approach to forecast long-term future human behavior in 3D, requiring only weak supervision from readily available 2D human action data.
This is a fundamental task enabling many downstream applications. The required ground-truth data is hard to capture in 3D (mocap suits, expensive setups) but easy to acquire in 2D (simple RGB cameras). 
Thus, we design our method to only require 2D RGB data at inference time while being able to generate 3D human motion sequences. We use a differentiable 2D projection scheme in an autoregressive manner for weak supervision, and an adversarial loss for 3D regularization.
Our method predicts long and complex human behavior sequences (e.g., cooking, assembly) consisting of multiple sub-actions. We tackle this in a semantically hierarchical manner, jointly predicting high-level coarse action labels together with their low-level fine-grained realizations as  characteristic 3D human poses.
We observe that these two action representations are coupled in nature, and joint prediction benefits both action and pose forecasting. 
Our experiments demonstrate the complementary nature of joint action and 3D pose prediction: our joint approach outperforms each task treated individually, enables robust longer-term sequence prediction, and improves over alternative approaches to forecast actions and characteristic 3D poses.

\end{abstract}
\vspace{-1em}

\section{Introduction}
\label{sec:introduction}
Predicting future human behavior is fundamental to machine intelligence, with many applications in content creation, robotics, mixed reality, and more.
For instance, a monitoring system might issue early warnings of potentially dangerous behaviour, and a robotic assistant can use forecasting to place tools at the right place and time they will be needed in the future.
Consider the specific scenario of an assembly line monitoring system deployed to issue early warnings of behavior that could be harmful in the near future: The system needs to have a long-term understanding of the future, enabling it to forecast multiple action steps ahead so that it can act in time before a harmful action occurs.  However, simply understanding the next action steps on a high level is not sufficient: it must also reason about \textit{where} the action will occur.   Actions such as ``grab a tool'' are likely harmless when performed in a toolbox; they become dangerous when done next to an active table saw or moving robot arm. The monitoring system thus also needs to be able to reason about spatial relations in 3D -- for both the location and body pose of involved humans.

To support these types of applications, we must address two tasks: 1) forecasting long-term action sequences, and 2) predicting future 3D human poses.
Prior work has focused on each of these tasks separately: activity forecasting predicts future action labels without considering the 3D poses  \cite{kitani2012activity,kuehne2014language,rhinehart2017first,furnari2020rolling,girdhar2021anticipative,DBLP:conf/cvpr/GongLKHC22}, while 3D pose forecasting focuses on fixed frame rate sequence prediction limited to single actions in short-term time frames without considering longer-term action sequences \cite{fragkiadaki2015recurrent,mao2019learning,mao2020history,yuan2020dlow,DBLP:conf/eccv/XuWG22}.

We propose that these two tasks are coupled in nature: predicting action labels with realized 3D poses helps to encourage richer feature learning and can materialize sub-category level differences in actions for predicting future activities, and grounding 3D poses with actions provides global structure for longer-term forecasting.   

Leveraging this insight,  we design a method that takes in a sequence of recent RGB image observations and their action labels, and jointly predicts a sequence of future 3D characteristic poses and action labels (Fig.~\ref{fig:teaser}).
In our design, we had to address two significant research challenges: 1) forecasting 3D poses from 2D images without any paired 3D training data, and 2) forecasting long sequences of actions comprising several discrete action steps.  

The first challenge arises from limited training data.   It would be ideal to have a dataset with ground truth 3D pose and action annotations for complex sequences of actions. Unfortunately, no such dataset exists. There are RGB video datasets with tracked 3D poses for limited types of actions (e.g., walking or waving); and there are video datasets with action labels for complex sequences of actions (e.g., cooking or assembly). However, there is no single dataset that has both types of annotations, and capturing one would be difficult due to the challenges of setting up 3D pose trackers in settings where people typically perform complex sequences of actions (e.g., cooking in a kitchen).  Instead, we have to learn to use 2D video observations for 3D pose and action label forecasting without paired data.
We achieve this by weakly supervising our pose forecasting in 2D using readily available 2D action datasets~\cite{rohrbach15ijcv,ben2021ikea} and formulate an adversarial loss encouraging likely 3D characteristic poses with respect to a distribution learned from 3D pose datasets~\cite{DBLP:journals/pami/IonescuPOS14,AMASS:ICCV:2019,GRAB:2020}. Crucially, this does not require any correspondence between the 2D video and 3D pose data.

The second challenge arises from the difficulties of predicting long sequences of discrete events.  One option would be to train a model to output a multi-step sequence of actions and poses all at once -- however, this is impossible given the exponential growth of multi-step sequences and the limited amount of available training data.  Another option would be to train a model that predicts the next future poses and actions at fixed time points in the future (e.g., 1s in advance) and then recurrently make long-term predictions -- however, this time-based forecasting approach produces sequences that tend to ``drift'' over the long-term, since the intermediate poses at fixed time steps are usually ``in between'' semantically meaningful actions and thus do not provide a distinctive input representation for the next prediction.  To address this issue, we train our autoregressive approach to iteratively generate the next discrete action label along with the \emph{3D characteristic pose} for that action.  A 3D characteristic pose \cite{diller2020forecasting} is the set of 3D joint positions corresponding to the most distinctive moment a semantic action is being performed (e.g., when a hand grasps an object, when two objects are first brought together, etc.).   By training our method to produce these poses as intermediate outputs (and inputs to the next step), we are able to generate more semantically plausible forecasts over longer action sequences.

Our experiments with two RGB video datasets demonstrate that our approach for joint prediction of action behaviors and 3D poses outperforms state-of-the-art methods applied separately to each task. 
Additionally, we find that predicting actions and their 3D characteristic poses enables more robust autoregressive prediction for longer-term forecasting.
Overall, our contributions are:
\begin{itemize}
    \item The first method to learn forecasting of future 3D poses from datasets with only 2D RGB video and action label data (i.e., without any paired 3D data).
    \item The first method to forecast future 3D poses jointly with action labels from commonly available video input.
    \item The first method to forecast future characteristic 3D poses and action labels for long-term and complex behaviors.
\end{itemize}

\section{Related Work}
\label{sec:related-work}

\begin{figure*}
\centering
\includegraphics[width=\textwidth]{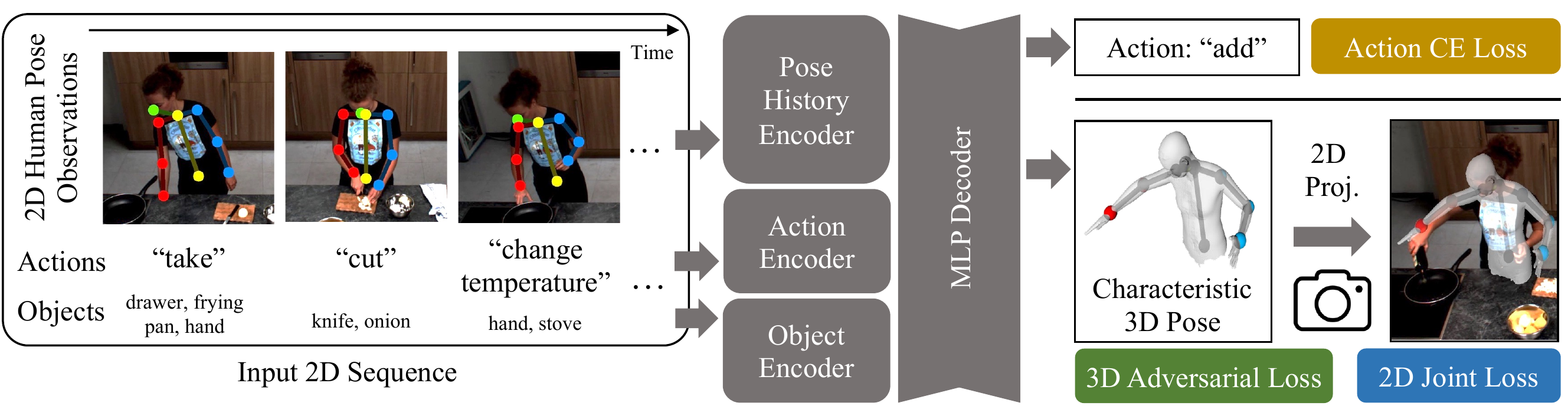}
\caption{Our approach takes as input a sequence of RGB images, from which 2D poses are extracted, as well as  their corresponding action label and initial set of objects.
Each input is encoded into a joint latent space to jointly predict the next action label and characteristic 3D pose.
While action labels are directly supervised, the 3D pose decoder is trained to match the next 2D action pose using differentiable projection, and an adversarial 3D loss encourages valid 3D pose prediction.
}
\label{fig:method-overview}
\end{figure*}

\noindent
\textbf{3D Human Pose Forecasting.}
Forecasting 3D human poses has been studied in many previous works and is commonly formulated as a 3D sequential motion prediction task, taking an input 3D sequence of poses and generating an output 3D sequence of poses.
For short-term future prediction (up to $\approx1$ second), RNN-based approaches have achieved impressive performance \cite{DBLP:conf/iccv/FragkiadakiLFM15,DBLP:conf/cvpr/JainZSS16,DBLP:conf/cvpr/MartinezB017,DBLP:conf/iccv/AksanKH19,DBLP:conf/wacv/ChiuAWHN19,DBLP:conf/iccv/WangACHN19,DBLP:conf/cvpr/GopalakrishnanM19,DBLP:conf/eccv/GuiWLM18,DBLP:conf/bmvc/PavlloGA18}.
As RNNs tend to struggle to capture longer-term dependencies with a fixed-size history, graph-based \cite{mao2019learning,DBLP:conf/iccv/DangNLZL21,DBLP:conf/iccvw/LiCLZXTZ21,DBLP:conf/cvpr/LiCZZW020,DBLP:conf/iccv/SofianosSFG21,DBLP:conf/cvpr/ZhongHZYX22,DBLP:conf/cvpr/CuiSY20,DBLP:conf/cvpr/CuiS21,DBLP:journals/tip/LiTZFL21,DBLP:conf/eccv/XuWG22} and attention-based \cite{DBLP:conf/ijcai/TangMLZ18,mao2020history,DBLP:conf/3dim/AksanKCH21,DBLP:conf/eccv/CaiHWC0YLYZSLLM20,DBLP:conf/iccvw/Martinez-Gonzalez21} approaches have been proposed to encode temporal history. Some methods also explored the applicability of temporal convolutions \cite{DBLP:conf/cvpr/LiZLL18,DBLP:conf/cvpr/MedjaouriD22} and MLP-only architectures \cite{DBLP:conf/wacv/GuoDSLAM23,DBLP:conf/ijcai/BouaziziHKDB22} to the task of human motion forecasting.
Additionally, various approaches have been proposed to model future human motion stochastically to produce diverse future sequence predictions, either with adversarial GAN formulations \cite{barsoum2018hp,DBLP:conf/aaai/KunduGB19}, conditional variational autoencoders (VAEs) \cite{yan2018mt,DBLP:conf/iccv/WalkerMGH17,aliakbarian2020stochastic,DBLP:conf/eccv/XuWG22,mao2021gsps,DBLP:conf/iccv/Cai0ZC0Y0ZYDSLM21,DBLP:conf/cvpr/SalzmannPR22,DBLP:conf/eccv/LucasBWR22,DBLP:conf/cvpr/BlattmannMDO21a}, or diverse sampling \cite{yuan2020dlow,DBLP:conf/mm/DangNLZL22}. More recently, diffusion methods \cite{DBLP:conf/icml/Sohl-DicksteinW15,DBLP:conf/iclr/SongME21} have been used for human motion generation and forecasting \cite{tanke2023social,xu2023interdiff,DBLP:conf/cvpr/JiangCPSZA23,DBLP:journals/corr/abs-2208-15001,DBLP:conf/iclr/TevetRGSCB23,DBLP:conf/cvpr/ZhouW23,DBLP:conf/cvpr/DabralMGT23,DBLP:conf/iccv/BarqueroEP23}.
These methods require 3D ground truth sequences for training, limiting applicability to scenarios where 3D inputs and ground-truth are not available. Ours requires only 2D training data for the action sequences, which is far more plentiful and easier to obtain. We generate valid 3D poses by leveraging an adversarial loss formulation, operating on a database of uncorrelated 3D poses.

\noindent
\textbf{Human Action Forecasting.}
Action forecasting has been studied by many approaches to predict future actions from a sequence of observed actions \cite{DBLP:conf/iccvw/FarhaG19,DBLP:conf/cvpr/FarhaRG18,DBLP:conf/cvpr/KeFS19,DBLP:conf/cvpr/FernandoH21} or directly from an input video sequence \cite{DBLP:conf/dagm/FarhaKSG20,DBLP:conf/eccv/SenerSY20,DBLP:conf/iccv/FurnariF19,DBLP:conf/iccv/GammulleDSF19,girdhar2021anticipative,DBLP:conf/cvpr/MiechLSWTT19,DBLP:conf/eccv/SenerSY20,DBLP:conf/iccv/SenerY19}.
Various methods have been developed to learn effective representations, including Hidden Markov Models \cite{kuehne2014language}, RNNs \cite{abu2018will,jain2016recurrent,wu2020learning,DBLP:conf/dagm/FarhaKSG20,DBLP:conf/eccv/SenerSY20,DBLP:conf/iccvw/FarhaG19,DBLP:conf/cvpr/FarhaRG18,DBLP:conf/iccv/FurnariF19}, transformer-based networks \cite{girdhar2021anticipative,DBLP:conf/cvpr/GongLKHC22,DBLP:journals/tip/RoyF21}, and self-supervised feature learning \cite{vondrick2016anticipating,han2020memory}.
There are approaches that focus on the short-term future \cite{DBLP:conf/cvpr/FernandoH21,DBLP:conf/iccv/FurnariF19,DBLP:conf/iccv/GammulleDSF19,girdhar2021anticipative,DBLP:conf/cvpr/MiechLSWTT19,DBLP:journals/tip/RoyF21,DBLP:conf/eccv/SenerSY20,DBLP:conf/iccv/SenerY19} or on longer-term actions \cite{DBLP:conf/cvpr/GongLKHC22,DBLP:conf/iccvw/FarhaG19,DBLP:conf/cvpr/FarhaRG18,DBLP:conf/cvpr/KeFS19,DBLP:conf/cvpr/FernandoH21,DBLP:conf/dagm/FarhaKSG20,DBLP:conf/eccv/SenerSY20,DBLP:conf/iccv/FurnariF19,DBLP:conf/iccv/GammulleDSF19,girdhar2021anticipative,DBLP:conf/cvpr/MiechLSWTT19,DBLP:conf/eccv/SenerSY20,DBLP:conf/iccv/SenerY19}.
Such method focus on characterizing anticipation with action labels only, while we aim to predict a richer characterization of the anticipated future by leveraging characteristic 3D poses, representative of future action goals in a sequence of action-pose predictions.
Forecasting actions alongside human poses in 2D only has been studied in a few works, for 2D hand placement~\cite{liu2020forecasting} or full-body 2D human poses at most 1 second into the future~\cite{zhu2021and}. 
Our approach addresses the benefits of 3D reasoning in human motion forecasting, without requiring full 3D sequences for supervision.

\noindent
\textbf{Goal-Driven Future Prediction.}
Goal-driven forecasting has previously been explored beyond action label forecasting, and has been leveraged to predict goal locations for future human walking trajectories \cite{cao2020long} and for future video sequences by predicting keyframes \cite{jayaraman2018time,bar2021compositional,pertsch2020keyframing,epstein2021learning}. Diller et al. \cite{diller2020forecasting} introduced the task of forecasting \textit{characteristic 3D poses}, salient keyframe poses representing the next action. These goal-based poses are more semantically meaningful and consistent across different action sequences than time-based ones, and thus are better suited for long-term forecasting.
We build upon these ideas by introducing a new goal-driven method for joint action anticipation and characteristic 3D pose forecasting in an auto-regressive system that can predict complex, long-term behavior sequences.

\section{Method Overview}
\label{sec:method-overview}

Our method aims to learn to jointly model future human actions along with the characteristic 3D poses representative of those actions.
From a sequence of RGB image observations of a person performing a series of actions and the corresponding action labels, we predict a sequence of future action labels and 3D poses characteristic of these actions.
This enables joint reasoning of not only global semantic behavior but also the physical manifestation thereof.

In the absence of 3D pose data of complex human actions, we weakly supervise forecasted 3D poses to align to future poses in 2D, and constrain the poses to be valid in 3D using an adversarial loss with a database of 3D poses. This does not require any correspondence between 3D pose data and 2D video, enabling action sequence supervision on commonly available 2D human action data together with carefully captured but unrelated human poses in 3D.

An overview of this approach is shown in Fig.~\ref{fig:method-overview}.
For an input sequence $S=\{(I_i,a_i,o_i)\}$ of $N$ RGB images $\{I_i\}$ with corresponding actions $\{a_i\}$ and initially involved objects $\{o_i\}$, we aim to predict the future $M$ actions $\{\hat{a}_k\}$ that will be taken along with their characteristic poses in 3D $\{\hat{Y}_k\}$.
We define the human pose as a collection of $J$ body joints at salient locations, so each output pose $\hat{Y}_k$ is predicted as a set of $J$ 3D coordinates.
We first extract information about the observed 2D pose movement by detecting 2D poses $\{X_i\}$, each with $J$ 2D joints, with a state-of-the-art 2D pose estimator that seamlessly integrates into our pipeline in a pre-trained and frozen form.

Next, we encode this information along with previously observed action and object labels to predict the next future action label $\hat{a}_k$ and characteristic 3D pose $\hat{Y}_k$.
We can then forecast a future sequence by autoregressively predicting a series, considering the 2D projections of the previously predicted 3D poses along with previously predicted actions as input to a new prediction.

\section{Joint Forecasting of Actions and Characteristic 3D Poses}
\label{sec:method-details}

Our network takes as input the previous 2D observations $\{X_i\}$ extracted from the $\{I_i\}$ images, as well as action and object labels $\{a_i\}$ and $\{o_i\}$ as one-hot vectors. 
Since we only predict action labels, object labels are given from the objects seen at the beginning of the sequence, and subsequently re-used for the entire sequence.
Each of these are encoded in parallel with three separate encoders; the actions and objects with MLPs while the poses are projected into latent space with a single linear layer and then processed with a stack of three residual blocks.
These encoded features are then all concatenated together in latent space, and processed jointly with an MLP to produce a common latent code $z$.
Finally, we decode both poses and actions in parallel based on $z$ using an MLP decoder each, yielding the next action label class as a vector $\hat{a}_k\in \mathbb{R}^{N_a}$ and 3D characteristic pose $\hat{Y}_k \in \mathbb{R}^{J\times 3}$, with $N_a$ the number of action classes. 
For a more detailed architecture specification, we refer to the appendix.

We jointly learn future action labels and characteristic 3D poses by supervising $\hat{a}_k$ and $\hat{Y}_k$ to match the observed future 2D video, and constrain $\hat{Y}_k$ to form a valid 3D pose by an adversarial loss, optimizing for the overall loss:
\begin{equation}
    \mathcal{L} = \lambda_{action}\mathcal{L}_{action} + \lambda_{pose2d}\mathcal{L}_{pose2d} + \lambda_{adv3d}\mathcal{L}_{adv3d}
\end{equation}
where $\mathcal{L}_{action}$ denotes the action loss, as described in Sec.~\ref{sec:method-details-actions}, $\mathcal{L}_{pose2d}$ and $\mathcal{L}_{adv3d}$ constraining the predicted pose, as described in Sec.~\ref{sec:method-details-poses}, and the $\lambda$ weighting each loss.

\subsection{Action Forecasting}
\label{sec:method-details-actions}
Predicted future actions are decoded from the latent code $z$ by an MLP decoder to predict the action class $\hat{a}_k$, supervised by cross entropy with the ground truth future action: $\mathcal{L}_{action}=\mathrm{CE}(\hat{a}_k, a^{\mathrm{gt}}_k)$.

\subsection{Characteristic Pose Forecasting}
\label{sec:method-details-poses}
Our goal is to forecast complex action behavior not only in terms of action labels, but also manifested as a sequence of characteristic poses in 3D.
Since we only have 2D pose annotations available, we first constrain these poses to represent future actions in 2D and make use of an adversarial regularization in 3D.
This does not require any correspondence between 2D and 3D data, only a collection of valid 3D poses, which are readily available.

\noindent
\textbf{Differentiable 2D Projection}
Our generator network predicts the next characteristic action pose $\hat{Y}_k$ as a set of 3D joints. 
To constrain $\hat{Y}_k$ based on the target future 2D pose $X^\mathrm{gt}$ extracted from the ground truth future image, we differentiably project $\hat{Y}_k$ into the 2D image with intrinsic parameters $K$ and extrinsic rotation and translation $R,t$:
\begin{equation}
    \hat{X} = K (R \hat{Y}_k + t)
\end{equation}

Since we learn from third-person video with a fixed camera, we can use the same camera parameters for all sequences used for training.
We can then define the 2D pose loss as the mean squared error between the projected pose prediction and the ground truth:
\begin{equation}
    \mathcal{L}_{pose2d} = ||X^\mathrm{gt} - X_k||^2_2
\end{equation}

Note that we only predict the $J$ joints that have been observed in the video data (excluding any joints that remain occluded in the observed video data), so this loss can be applied to all predicted joints.

\smallskip
\noindent
\textbf{Adversarial 3D Pose Regularization.}
While the action and pose prediction losses provide effective predictions when considered in the 2D projections, the $\{\hat{Y}_k\}$ remain underconstrained in 3D and  thus tend to exhibit large distortions and implausible bone lengths and angles, when trained with only 2D supervision.
We thus constrain the predicted poses to form valid 3D poses by formulating an adversarial 3D loss from a  critic network which is simultaneously trained to distinguish predicted poses from a database of real 3D skeleton samples. 
Note that there is no correspondence between these skeletons and the 2D poses extracted from the action video sequences -- any database of 3D skeletons can be used. 
We can thus train our approach with an entirely uncorrelated 3D pose dataset without requiring 3D action pose correlations.

We then formulate $\mathcal{L}_{adv3d}$ as a Wasserstein loss \cite{arjovsky2017wasserstein}, training the critic network in an alternating fashion with  the generator. 
This enables effective forecasting of future 3D characteristic poses for predicted future action labels, without requiring any 3D observations as input.

In order to enable the critic network to learn effectively about likely intrinsic pose constraints (e.g., lengths, kinematic chains, or valid joint angles), the critic takes as input not only the 3D joint locations of $\hat{Y}_k$ but also their kinematic statistics as a matrix $\Psi$, following \cite{wandt20163d,wandt2019repnet}.

$\Psi$ encodes joint angles and bone lengths as $\Psi=B^TB$, where $B=(b_1, b_2, \dots, b_b)$ is a matrix with columns $b_i=j_k-j_l$ representing the vectors between each joint $j_k$ and $j_l$.
$\Psi$ then contains bone  lengths $l_i^2$ on its diagonal, and angular representations on the off-diagonal entries.

\begin{table*}[!t]
\begin{center}
\resizebox{0.9\textwidth}{!}{\begin{tabular}{|l||c|c||c|c||c|c||c|c|}
\hline
 & \multicolumn{4}{c||}{MPII Cooking II} & \multicolumn{4}{c|}{IKEA ASM} \\
\hline
 & 2d & 3d & \multicolumn{2}{c||}{Action Accuracy} & 2d & 3d & \multicolumn{2}{c|}{Action Accuracy} \\
\hline
Approach & MPJPE [px] $\downarrow$ & Quality $\uparrow$ & top-1 $\uparrow$ & top-3 $\uparrow$ & MPJPE [px] $\downarrow$ & Quality $\uparrow$ & top-1 $\uparrow$ & top-3 $\uparrow$\\
\hline\hline
Zero Velocity & 118 & -- & -- & -- & 74 & -- & -- & -- \\
Train Average & 166 & -- & -- & -- & 91 & -- & -- & -- \\
\hline\hline
AVT \cite{girdhar2021anticipative} RGB & -- & -- & 19\% & 42\% & -- & -- & 22\% & 49\% \\
AVT \cite{girdhar2021anticipative} RGB+Skeleton & -- & -- & 20\% & 40\% & -- & -- & 23\% & 47\% \\
FUTR \cite{DBLP:conf/cvpr/GongLKHC22} RGB & -- & -- & 27\% & 48\% & -- & -- & 19\% & 45\% \\
FUTR \cite{DBLP:conf/cvpr/GongLKHC22} RGB+Skeleton & -- & -- & 27\% & 49\% & -- & -- & 20\% & 46\% \\
\hline\hline
RepNet \cite{wandt2019repnet} + DLow (min-10) \cite{yuan2020dlow} & 72 & \textbf{0.72} & -- & -- & 45 & 0.31 & -- & -- \\
RepNet \cite{wandt2019repnet} + GSPS (min-10) \cite{mao2021gsps} & 59 & 0.66 & -- & -- & 51 & 0.15 & -- & -- \\
RepNet \cite{wandt2019repnet} + STARS (det.) \cite{DBLP:conf/eccv/XuWG22}  & 70 & 0.62 & -- & -- & 54 & 0.27 & -- & -- \\
RepNet \cite{wandt2019repnet} + EqMotion \cite{xu2023eqmotion}  & 68 & 0.66 & -- & -- & 55 & 0.23 & -- & -- \\
\hline\hline
Joint 2D Pose \& Action \cite{zhu2021and} & 55 & - & 27\% & 43\% & 44 & - & 22\% & 46\% \\
\hline\hline
\textbf{Ours} & \textbf{50} & 0.55 & \textbf{29\%} & \textbf{51\%} & \textbf{40} & \textbf{0.31} & \textbf{29\%} & \textbf{50\%}  \\
\hline
\end{tabular}}
\vspace{-0.5em}
\caption{
Quantitative comparison with state-of-the-art action label and 3D pose forecasting. Our joint approach enables more accurate future action and pose predictions, compared to approaching both tasks separately, and outperforms joint action and 2D pose forecasting.
}
\label{tab:baselines}
\end{center}
\vspace{-2.5em}
\end{table*}

\subsection{Sequence Prediction}
\label{sec:method-details-sequence}
In order to forecast longer-term future behavior, our 3D pose predictions enable a natural autoregressive sequence prediction by taking the predictions $\hat{X}_t, \hat{a}_t$ at time step $t$ as part of the input for time step $t+1$.
We can thus predict a sequence of $M$ future action labels $\{\hat{a}_k\}$ and characteristic 3D poses $\{\hat{Y}_k\}$; we use $M=10$ for MPII Cooking II~\cite{rohrbach15ijcv} and $M=5$ for IKEA-ASM~\cite{ben2021ikea}, respectively.

\subsection{Training Details}
\label{sec:method-details-training}

We train our approach for the $J=9$ joints commonly seen across the input observed video data, characterizing the upper body in MPII Cooking II~\cite{rohrbach15ijcv} and IKEA-ASM~\cite{ben2021ikea}. 

Additionally, we use loss weights $\lambda_{action} = 1e^6$, $\lambda_{pose} = 1$, and $\lambda_{adv3d} = 1$, empirically chosen to numerically balance each individual loss with the others.

We train our approach on a single NVIDIA GeForce RTX 2080TI for $\approx 12$ hours until convergence. We use ADAM with batch size 4096, weight decay  0.001, and a constant learning rate of 0.0001 for both generator and discriminator. 

\subsection{Datasets}
\label{sec:method-details-data}

We train and evaluate our approach on two datasets: MPII Cooking II~\cite{rohrbach15ijcv} and IKEA-ASM~\cite{ben2021ikea}. Both datasets contain sequences of human actors performing complex, unscripted actions, and provide annotations of fine-grained sub-action steps.
MPII Cooking II~\cite{rohrbach15ijcv} is an action recognition dataset with 272 complex cooking sequences and an average sequence time of 182s (35 annotated sub-actions, each 5.2s on average). 
IKEA-ASM contains 370 sequences of actors assembling IKEA furniture, with an average of 74s per sequence (15 annotated sub-actions, each 4.9s on average).

In both datasets, each action sequence has been filmed from a fixed camera setup; the third-person point of view enables extraction of 2D poses with an off-the-shelf 2D pose estimator.
We use OpenPose~\cite{cao19OpenPose} in our experiments and note that our approach is agnostic to the concrete method of 2D pose detection. We provide more in-depth discussion and additional experiments in the appendix.

We consider the 9 upper-body joints of the OpenPose skeletons, as the other joints are almost always occluded in the video observations, and remove global translation by centering each 2D pose at the neck joint.

Characteristic poses, in contrast to an arbitrary pose within a labeled action range, are the most representative pose of that action, and are annotated for all sub-actions in each sequence as the most articulated pose of that sub-action, following the annotation protocol of \cite{diller2020forecasting}.
Annotation can be done efficiently and was performed by the authors within just 32 hours, yielding a total of $\approx$18,000 characteristic poses ($\approx$12,000 for MPII Cooking II and $\approx$6,000 for IKEA-ASM). 
These poses are indicative of the action they represent as demonstrated in Tab. \ref{tab:ablations-charposes}: Using such poses significantly improves performance, validating our annotation protocol.

For the 3D adversarial loss, we use $\approx$800,000 human poses from popular 3D pose datasets: Human3.6m~\cite{DBLP:journals/pami/IonescuPOS14}, AMASS~\cite{AMASS:ICCV:2019}, and GRAB~\cite{GRAB:2020}. 
Note that none of these 3D poses have any correspondence with the 2D posed actions from the MPII Cooking II dataset, instead depicting various human skeletons in natural and diverse poses.

\section{Results}
\label{sec:evaluation}

We evaluate sequence forecasting of action labels and characteristic 3D poses on the MPI Cooking II~\cite{rohrbach15ijcv} and IKEA-ASM~\cite{ben2021ikea} datasets, and 3D pose quality by comparing to our database of high-fidelity 3D poses.

\subsection{Evaluation Metrics}
\label{sec:evaluation-metrics}
\noindent
\textbf{2D Pose Error.}
Since we only have 2D ground-truth data available for complex action sequences, we first project predicted 3D poses back into 2D, and evaluate the 2D mean per-joint position error (MPJPE)~\cite{DBLP:journals/pami/IonescuPOS14}, in comparison with 2D poses extracted from ground-truth future frames using \cite{cao19OpenPose}:
$\text{E}_{\text{MPJPE}} = \frac{1}{M}\sum_{j=1}^{M} ||\hat{X} - X^\mathrm{gt}||_2$.

\begin{figure}[t]
\centering
\includegraphics[width=0.85\columnwidth]{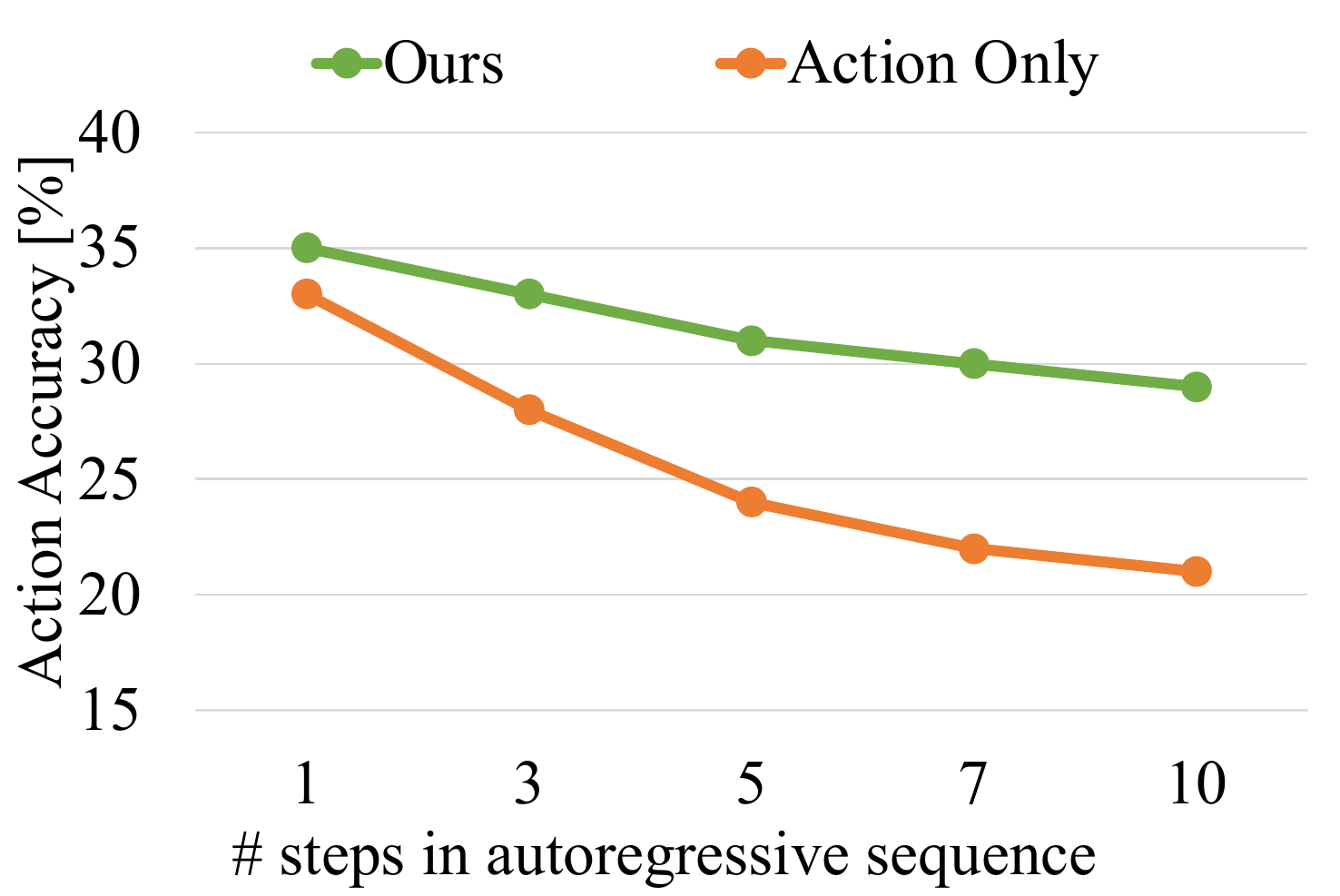}
\vspace{-0.5em}
\caption{Action accuracy over time. Our joint action-characteristic pose forecasting enables more robust autoregressive action forecasting than action prediction without considering pose.}
\label{fig:action-accuracy-lines}
\vspace{-1.5em}
\end{figure}

\noindent
\textbf{3D Pose Quality.} 
In the absence of annotated ground truth 3D poses for the action video sequences, we measure the quality of predicted 3D poses as how distinguishable they are in comparison to a set of real 3D poses.
We follow \cite{aliakbarian2020stochastic} and evaluate quality by training a binary classifier on ~50,000 human poses generated at different training steps (representing examples of unrealistic 3D poses) and ~50,000 real 3D pose samples.
For classification accuracy $a$ of this classifier, quality is measured as $1-a$, with a quality of 1 indicating full indistinguishability from real poses. We refer to the appendix for more details on this quality metric.

\noindent
\textbf{Action Accuracy.}
We report the action accuracy of the predicted sequences, as the mean over all sequences in the test set.
We evaluate the top-$n$ accuracy based on whether the ground truth action is among the $n$ highest scoring predictions, for $n=1$ and $n=3$.

\subsection{Comparison to Human Pose Forecasting}
Tab.~\ref{tab:baselines} compares our method to state-of-the-art 3D pose forecasting methods DLow~\cite{yuan2020dlow}, GSPS~\cite{mao2021gsps}, STARS~\cite{DBLP:conf/eccv/XuWG22}, and EqMotion~\cite{xu2023eqmotion}.
These methods expect sequences of observed 3D human poses as input; we thus first apply a state-of-the-art weakly supervised 3D pose estimator~\cite{wandt2019repnet} on our 2D input poses, producing inputs and supervision in 3D. This method estimates 3D poses using an adversarial formulation, requiring a database of 3D poses not correlated with the 2D pose inputs. 
To ensure a fair comparison, this database is exactly the same as the one our method uses.

We chose the 3D pose estimator of \cite{wandt2019repnet} since its weakly supervised formulation is most comparable to our approach. An additional comparison to a fully supervised approach for 3D pose lifting (SPIN~\cite{kolotouros2019spin}) is provided in the appendix.

We then train the 3D pose prediction methods from scratch on this generated data, using their original parameter settings. Stochastic methods DLow and GSPS are set to predict 10 possible future sequences; we report the minimum error across these. We use STARS in the method's deterministic mode.
Each method takes as input a pose history of $M$ poses and outputs a sequence of $M$ poses, analogous to our setup where each pose is a characteristic pose corresponding to an action step ($M=10$ for MPII Cooking II and $M=5$ for IKEA-ASM).
Our approach to lift 2D to future 3D poses and actions in an end-to-end fashion enables more effective pose forecasting than these state-of-the-art 3D pose forecasting approaches on both datasets.

In addition, we compare to the joint 2D action and pose forecasting approach of Zhu et al. \cite{zhu2021and}. Our approach of forecasting long-term sequences of 3D poses alongside actions is able to outperform their 2D MPJPE pose prediction and action accuracy performance, due to improved spatial reasoning when forecasting 3D poses.

\noindent
\textbf{Statistical 2D Baselines.}
We additionally compare with two statistical baselines in 2D, following \cite{diller2020forecasting}: the average target train pose, and a zero-velocity baseline which was introduced by Martinez et al. \cite{DBLP:conf/cvpr/MartinezB017} as competitive with state of the art. We outperform both baselines, indicating that our method learns a strong action pose representation.

\subsection{Comparison to Action Label Forecasting}
We compare the action accuracy of our joint action-pose forecasting to AVT~\cite{girdhar2021anticipative} and FUTR~\cite{DBLP:conf/cvpr/GongLKHC22}, two state-of-the-art action anticipation methods, in Tab.~\ref{tab:baselines}.
We train and evaluate both AVT and FUTR on input RGB frames and their action and object labels, equal to our training setup, and use their original training settings initialized with a pre-trained vision transformer~\cite{dosovitskiy2020vit} for AVT and extracted I3D features \cite{DBLP:conf/cvpr/CarreiraZ17} from our datasets for FUTR. 
Additionally, as we consider extracted 2D poses from the input RGB images, we also evaluate a variant that is trained and evaluated on RGB images overlaid with 2D poses (``+Skeleton'').
Our approach outperforms these baselines in both scenarios, by jointly predicting future actions and characteristic 3D poses.

\begin{table}[b]
\vspace{-1em}
\begin{center}
\resizebox{\columnwidth}{!}{\begin{tabular}{|c|c||c||c|c||c|c|}
\hline
\multicolumn{2}{|c|}{Poses} & 2D & 3D & \multicolumn{2}{c|}{Action Accuracy} \\
\hline
Train & Test & MPJPE [px] $\downarrow$ & Quality $\uparrow$ & top-1 $\uparrow$ & top-3 $\uparrow$\\
\hline\hline
Uncoupled & Uncoupled & 75 & 0.29 & 28\% & 48\% \\
\hline
Middle & Middle & 58 & 0.45 & 26\% & 43\% \\
Random & Random & 67 & 0.37 & 22\% & 42\% \\
\textbf{Characteristic} & \textbf{Characteristic} & \textbf{50} & \textbf{0.55} & \textbf{29\%} & \textbf{51\%} \\
\hline
\end{tabular}}
\caption{
Ablation on pose forecasting on MPII Cooking II~\cite{rohrbach15ijcv}. Our characteristic pose representation maximizes MPJPE and action performance: We consider pose prediction following state-of-the-art pose forecasting as decoupled from actions (uncoupled), as well as poses coupled to actions but in the middle of an action range, or at a random time therein, and our characteristic pose prediction. The same pose type is used for both train and evaluation.
}
\vspace{-2em}
\label{tab:ablations-charposes}
\end{center}
\end{table}

\subsection{Ablation Studies}

\noindent
\textbf{What is the effect of pose forecasting on long-term action understanding?}
Tab.~\ref{tab:ablations} shows that there is a notable improvement in action accuracy between training only with an action loss vs. training action and 2D pose loss jointly. This becomes more apparent when training action only vs action and full pose prediction (2D and 3D losses). 
In addition, Fig.~\ref{fig:action-accuracy-lines} shows the correspondence between autoregressive prediction length and action accuracy: jointly forecasting poses and actions enables more robust autoregressive forecasting over time. We conclude that pose forecasting is beneficial for long-term action understanding.

\noindent
\textbf{How does action forecasting affect pose prediction performance?}
Tab.~\ref{tab:ablations} demonstrates that pose forecasting trained jointly with action prediction is complementary and enables more accurate pose prediction.

\begin{table*}[t]
\begin{center}
\resizebox{0.95\textwidth}{!}{\begin{tabular}{|c|c|c||c|c||c|c||c|c||c|c|}
\hline
\multicolumn{3}{|c||}{} & \multicolumn{4}{c||}{MPII Cooking II} & \multicolumn{4}{c|}{IKEA ASM} \\
\hline
\multicolumn{3}{|c||}{Losses During Training} & 2D & 3D & \multicolumn{2}{c||}{Action Accuracy} & 2D & 3D & \multicolumn{2}{c|}{Action Accuracy} \\
\hline
Action & 2D Proj. & 3D Adv. & MPJPE [px] $\downarrow$ & Quality $\uparrow$ & top-1 $\uparrow$ & top-3 $\uparrow$ & MPJPE [px] $\downarrow$ & Quality $\uparrow$ & top-1 $\uparrow$ & top-3 $\uparrow$ \\
\hline\hline
\checkmark & $\times$ & $\times$ & -- & -- & 21\% & 41\%  & -- & -- & 24\% & 45\% \\
\checkmark & $\checkmark$ & $\times$ & 62 & 0.10 & 26\% & 49\% & 46 & 0.05 & 27\% & 49\% \\
\hline
$\times$ & \checkmark & $\times$ & 54 & 0.21 & -- & -- & 44 & 0.09 & -- & --  \\
$\times$ & \checkmark & \checkmark & 58 & 0.53 & -- & -- & 43 & 0.29 & -- & --  \\
\hline\hline
\textbf{\checkmark} & \textbf{\checkmark} & \textbf{\checkmark} & \textbf{50} & \textbf{0.55} & \textbf{29\%} & \textbf{51\%} & \textbf{40} & \textbf{0.31} & \textbf{29\%} & \textbf{50\%}  \\
\hline
\end{tabular}}
\vspace{-0.5em}
\caption{Ablation on the effect of the action, 2D projection, and 3D adversarial losses. Combining all together for joint forecasting enables complementary learning to produce the best performance.}
\label{tab:ablations}
\end{center}
\vspace{-2.5em}
\end{table*}

\begin{figure*}[b]
\vspace{-1em}
\centering
\includegraphics[width=0.9\textwidth]{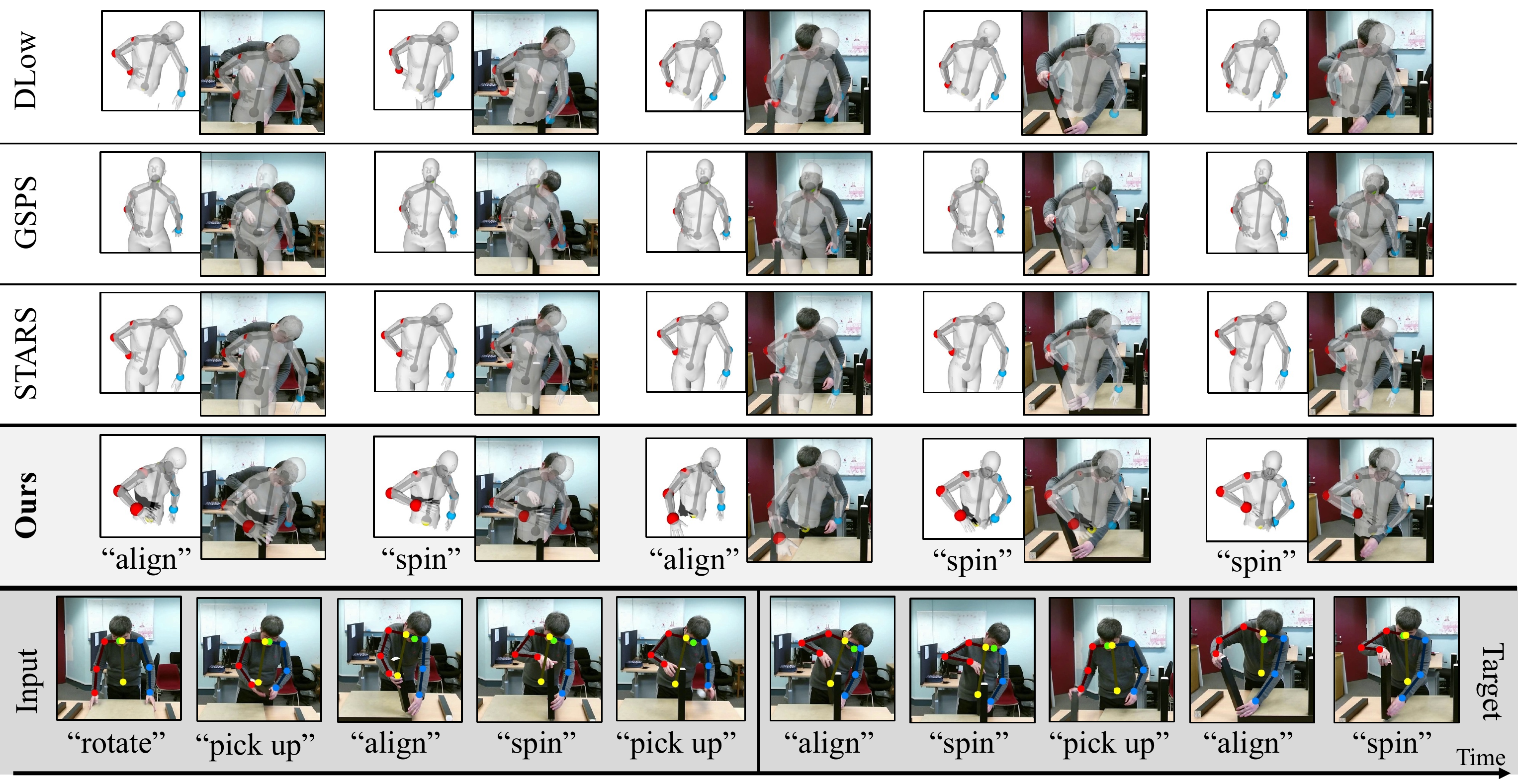}
\caption{Qualitative comparison between DLow \cite{yuan2020dlow}, GSPS \cite{mao2021gsps}, STARS \cite{DBLP:conf/eccv/XuWG22}, and our method on IKEA-ASM~\cite{ben2021ikea} data. For each method, we show the 3D predicted pose projected into the 2D target view, without background (small) and with background for context (full size). Our joint reasoning captures the individual characteristic action poses more faithfully while producing spatially plausible 3D poses.}
\label{fig:qualitative-ikea}
\vspace{-2em}
\end{figure*}

\noindent
\textbf{What is the effect of characteristic pose forecasting?}
Since state-of-the-art pose forecasting focuses on fixed frame rate predictions independent of actions, we compare with such joint forecasting of action and pose where predicted poses are sampled at equally spaced points in time in Tab.~\ref{tab:ablations-charposes} (uncoupled).
Additionally, we consider alternative poses to forecast for each action rather than a characteristic 3D pose (middle of the annotated action range, and randomly selected within the action range). We keep the same pose representation for training and testing (i.e., evaluate on middle poses when trained on them, etc.), for a fair comparison.
We observe the best performance when forecasting characteristic 3D poses along with action labels, showing their usefulness for forecasting long sequences of 3D poses and actions.

\subsection{Qualitative Results}
\begin{figure*}[b]
\vspace{-1em}
\centering
\includegraphics[width=0.95\textwidth]{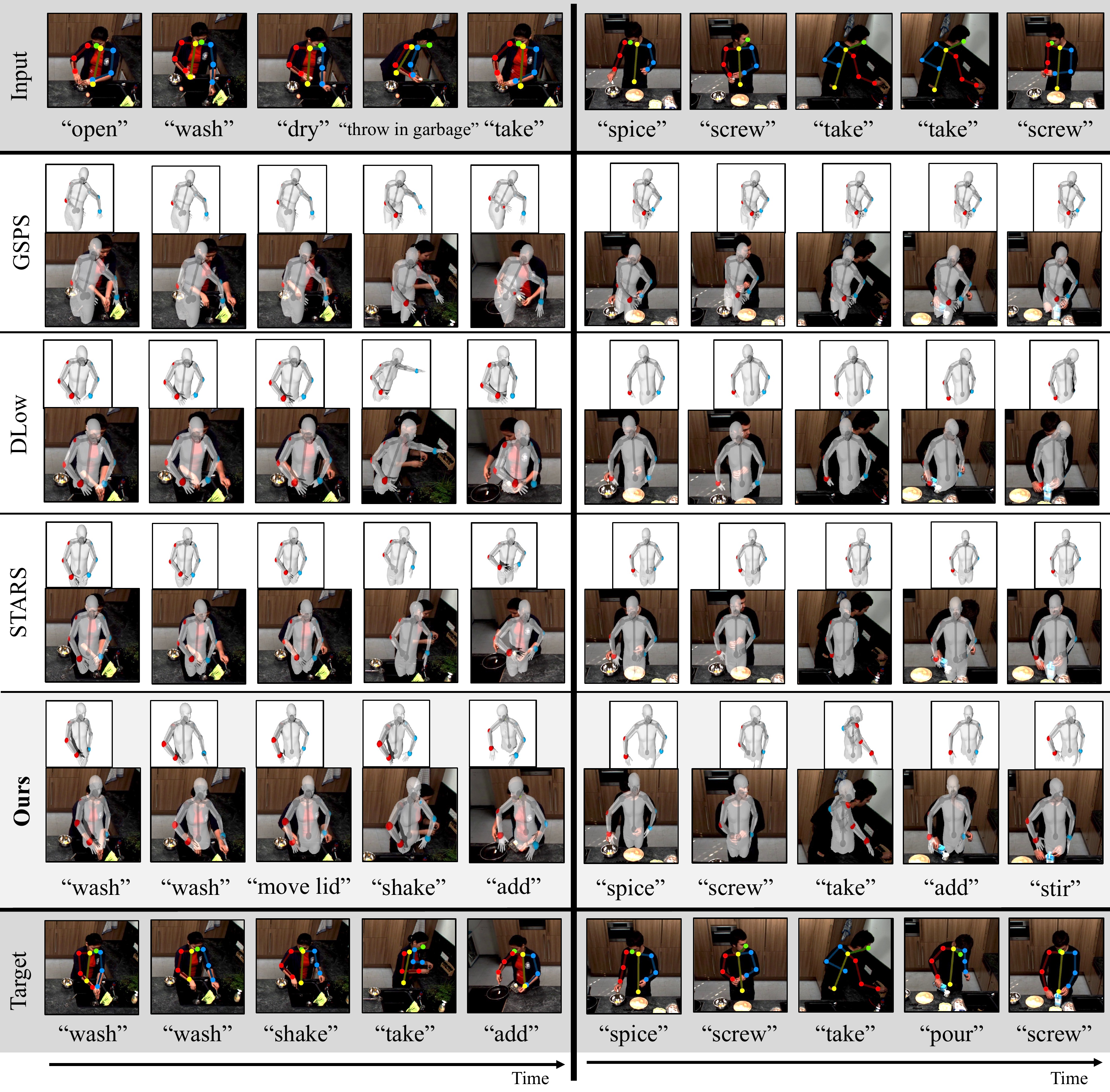}
\caption{Qualitative comparison between DLow \cite{yuan2020dlow}, GSPS \cite{mao2021gsps}, STARS \cite{DBLP:conf/eccv/XuWG22}, and our method on two sequences (left and right) from MPII Cooking II~\cite{rohrbach15ijcv}. For each method, we show the 3D predicted pose projected into 2D, without background (small) and with background for context (full size). By considering both 3D pose and action forecasting together, we more effectively forecast the longer-term behavior.}
\label{fig:qualitative}
\vspace{-1em}
\end{figure*}

Qualitative evaluations for the predicted poses are shown in Fig.~\ref{fig:qualitative} on data from MPII Cooking II~\cite{rohrbach15ijcv} and in Fig.~\ref{fig:qualitative-ikea} on data from IKEA-ASM~\cite{ben2021ikea}. We compare our approach with state-of-the-art 3D pose forecasting of DLow \cite{yuan2020dlow}, GSPS \cite{mao2021gsps}, and STARS \cite{DBLP:conf/eccv/XuWG22}.
For each method, we show a 3D body mesh in addition to the predicted 3D pose joints, to more comprehensively show the 3D structure of the forecasting results; we obtain body meshes by fitting SMPL~\cite{SMPL:2015} to each methods' predicted 3D body joints.

As there is no 3D ground truth available, we show the camera perspective with background for context as well as without background for a 3D pose only version.
The two views demonstrate the fit to the ground truth 2D along with the quality of the 3D pose, respectively. 
Our approach leads to poses that better follow the ground-truth action poses in 2D compared to both previous methods while still maintaining a valid pose structure in 3D.
Notably, this is true for both datasets, as our approach effectively forecasts the different data characteristics of both cooking as well as furniture assembly.
In particular, our joint action-3D pose forecasting enables more accurate forecasting with diverse and accurate 3D pose structures.

\subsection{Limitations}
While we have demonstrated the potential of joint action and 3D pose forecasting, several limitations remain.
For instance, our method leverages a separate 2D pose extraction as input to training, while an end-to-end formulation could potentially better leverage other useful signal in the input frames.
Additionally, a more holistic body representation than pose joints would be important for finer-grained interactions that involve reasoning over small limbs (e.g., hands) and body surface contact.

\vspace{-0.5em}
\section{Conclusion}
\label{sec:conclusion}

\vspace{-0.5em}

In this paper, we proposed to forecast future human behavior by jointly predicting future actions alongside characteristic 3D poses.
We do not require any 3D annotated action sequences, or 3D input data; instead, we learn complex action sequences from 2D action video data, and regularize predicted poses with an adversarial formulation against uncorrelated 3D pose data.
Experiments demonstrate that our joint forecasting enables complementary feature learning, outperforming each individual task considered separately.

\vspace{-0.5em}
\section*{Acknowledgements}
\label{sec:acknowledgements}

This project is funded by the Bavarian State Ministry of Science and the Arts and coordinated by the Bavarian Research Institute for Digital Transformation (bidt), and the German Research Foundation (DFG) Grant ``Learning How to Interact with Scenes through Part-Based Understanding".

{
    \small
    \bibliographystyle{ieeenat_fullname}
    \bibliography{main}
}

\clearpage
\renewcommand{\appendixpagename}{\Large{Appendix}} 

\begin{appendices}
We show in this appendix additional qualitative (Sec.~\ref{sec:additional-qualitative-results}) and quantitative (Sec.~\ref{sec:additional-quantitative-results}) results, detail our baseline evaluation protocol (Sec.~\ref{sec:baseline-evaluation}), elaborate on the 3D quality metric (Sec.~\ref{sec:quality-metric}), demonstrate the ability of our method to generalize to multi-actor scenarios (Sec.~\ref{sec:multi-actor}), verify our method's robustness to 2D detection results (Sec.~\ref{sec:2d-detectors}), show the architecture used in our approach (Sec.~\ref{sec:architecture}), and provide additional details regarding the data (Sec.~\ref{sec:data-details}).
\section{Additional Qualitative Results}
\label{sec:additional-qualitative-results}
\begin{figure*}[h]
\centering
\includegraphics[width=\textwidth]{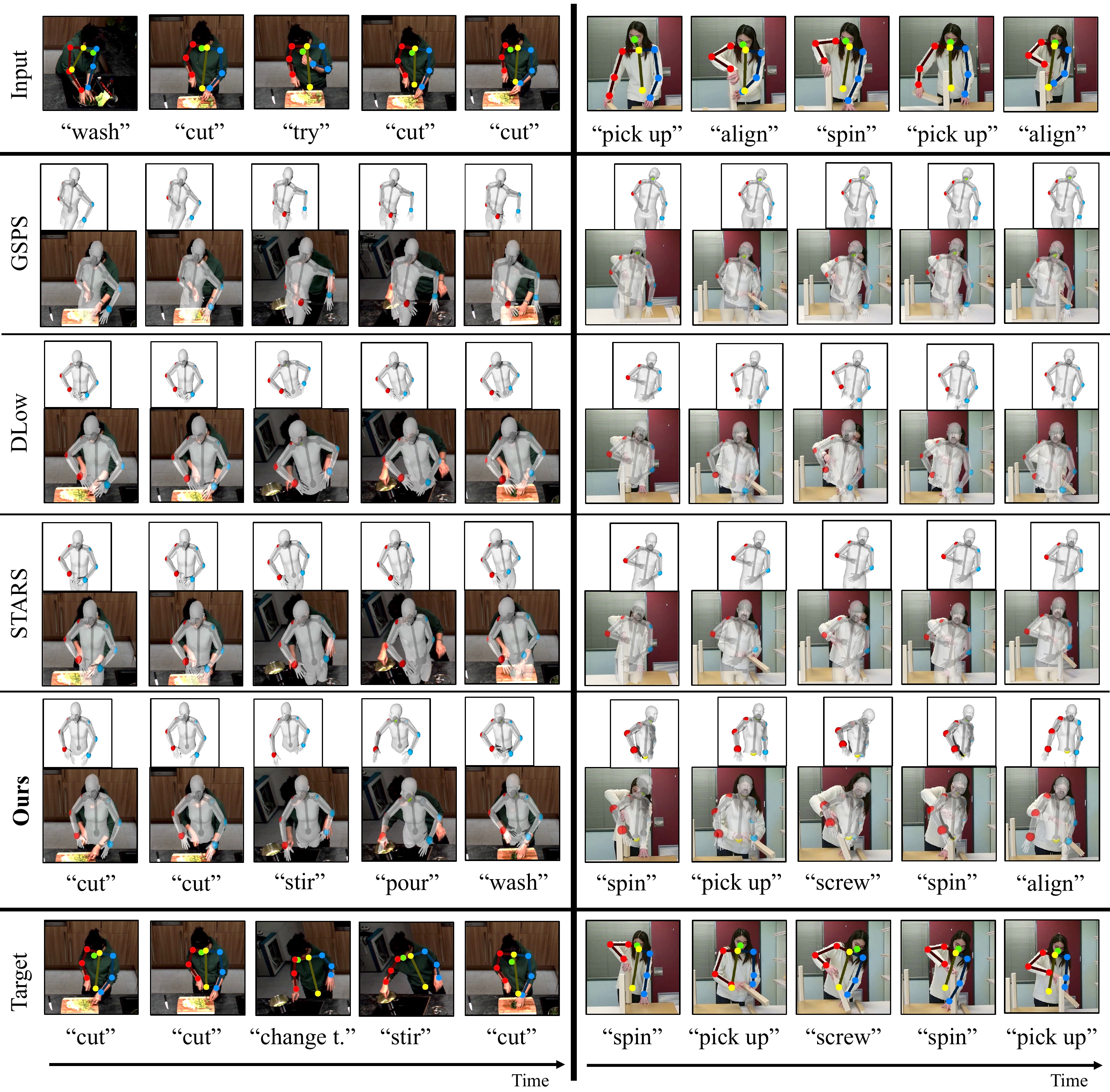}
\caption{Additional qualitative comparison between DLow \cite{yuan2020dlow}, GSPS \cite{mao2021gsps}, STARS \cite{DBLP:conf/eccv/XuWG22}, and our method on two sequences (left on MPII Cooking 2~\cite{rohrbach15ijcv}, right on IKEA-ASM~\cite{ben2021ikea}). For each method, we show a the 3D predicted pose projected into the 2D target view, without background for a pose only version (small) as well as with background for context (full size).}
\label{fig:qualitative-additional}
\end{figure*}

Fig.~\ref{fig:qualitative-additional} shows additional qualitative results of our method, on both MPII Cooking 2~\cite{rohrbach15ijcv} (left column) and IKEA-ASM~\cite{ben2021ikea} (right column), as compared to pose baselines DLow~\cite{yuan2020dlow}, GSPS~\cite{mao2021gsps}, and STARS~\cite{DBLP:conf/eccv/XuWG22}.

\section{Additional Quantitative Results}
\label{sec:additional-quantitative-results}
\subsection{Characteristic Poses}
Analogous to Tab.~2 in the main paper, Tab.~\ref{tab:ablations-charposes-ikea} shows an ablation on pose timings and compares our approach of using characteristic poses to poses taken at regular time intervals (``uncoupled'') as well as in the middle or at a random time of an action, on IKEA-ASM~\cite{ben2021ikea} data.
To further illustrate this point, Tab.~\ref{tab:ablations-charposes-density} shows additional ablations: Poses predicted at random points in the sequence, but at most 1s from the closest characteristic pose (``centered on the characteristic pose'') and predicting characteristic poses but evaluating interpolated regularly spaced poses. Both demonstrate that the usage of characteristic poses improves performance compared to other approaches while still being outperformed by directly predicting characteristic poses.
\begin{table}[b]
\begin{center}
\resizebox{\columnwidth}{!}{\begin{tabular}{|c|c||c|c||c|c|}
\hline
 & 2D & 3D & \multicolumn{2}{c|}{Action Accuracy} \\
\hline
Poses & MPJPE [px] $\downarrow$ & Quality $\uparrow$ & top-1 $\uparrow$ & top-3 $\uparrow$\\
\hline\hline
Uncoupled & 75 & 0.29 & 28\% & 48\% \\
\hline
Middle & 58 & 0.45 & 26\% & 43\% \\
Random & 67 & 0.37 & 22\% & 42\% \\
\hline
Centered on Char. Poses & 69 & 0.33 & 28\% & 50\% \\
Interp. from Char. Poses & 62 & 0.13 & 29\% & 51\% \\
\hline
\textbf{Characteristic} & \textbf{50} & \textbf{0.55} & \textbf{29\%} & \textbf{51\%} \\
\hline
\end{tabular}}
\caption{Ablation on pose forecasting on MPII Cooking II~\cite{rohrbach15ijcv}. We consider pose prediction following state-of-the-art pose forecasting as decoupled from actions (uncoupled), as well as poses coupled to actions in various fashions: middle (the middle pose of an action range), random (a random pose of the action), random but at most 1s from the closest characteristic pose (centered), regularly spaced poses interpolated from characteristic pose prediction, and our characteristic pose prediction.}
\label{tab:ablations-charposes-density}
\end{center}
\end{table}

\subsection{Lifting 2D Predictions to 3D}
In Tab.~1 in the main paper, we compare to first lifting input poses into 3D, then performing 3D motion prediction. Tab.~\ref{tab:2d-lift-3d} evaluates the other way around: Predicting 2D poses and action labels jointly with \cite{zhu2021and}, then lifting the predicted 2D poses into 3D with RepNet \cite{wandt2019repnet} for evaluation. Our method outperforms both approaches.

\begin{table}[h]
\vspace{-1em}
\begin{center}
\resizebox{\columnwidth}{!}{\begin{tabular}{|l||c|c||c|c||c|c||c|c|}
\hline
 & \multicolumn{4}{c||}{MPII Cooking II} & \multicolumn{4}{c|}{IKEA ASM} \\
\hline
 & 2d & 3d & \multicolumn{2}{c||}{Action Accuracy} & 2d & 3d & \multicolumn{2}{c|}{Action Accuracy} \\
\hline
Approach & MPJPE [px] $\downarrow$ & Quality $\uparrow$ & top-1 $\uparrow$ & top-3 $\uparrow$ & MPJPE [px] $\downarrow$ & Quality $\uparrow$ & top-1 $\uparrow$ & top-3 $\uparrow$\\
\hline\hline
\cite{zhu2021and} + \cite{wandt2019repnet} & 63 & 0.50 & 27\% & 43\% & 53 & 0.21 & 22\% & 46\% \\
\textbf{Ours} & \textbf{50} & \textbf{0.55} & \textbf{29\%} & \textbf{51\%} & \textbf{40} & \textbf{0.31} & \textbf{29\%} & \textbf{50\%} \\
\hline
\end{tabular}}
\caption{Our approach of jointly forecasting 3D poses and actions achieves better performance compared to 2D pose + action forecasting \cite{zhu2021and} and then lifting forecasted 2D poses into 3D using \cite{wandt2019repnet}.}
\vspace{-2em}
\label{tab:2d-lift-3d}
\end{center}
\end{table}

\subsection{Input Noise Ablation}
Tab.~\ref{tab:nonoise-noobjects} shows the effect using a noise vector as additional input to our method. It encourages more diversity in predictions, which benefits pose and action forecasting.

\subsection{Input Objects Ablation}
Inputting initially observed objects slightly improves results (Tab.~\ref{tab:nonoise-noobjects}), due to added context for broad actions like ``add,'' e.g.``add ingredient'' vs. ``add water to pot.''.

\begin{table}[h]
\begin{center}
\resizebox{\columnwidth}{!}{%
\begin{tabular}{|l||c|c||c|c||c|c||c|c|}
\hline
 & \multicolumn{4}{c||}{MPII Cooking II} & \multicolumn{4}{c|}{IKEA ASM} \\
\hline
 & 2d & 3d & \multicolumn{2}{c||}{Action Accuracy} & 2d & 3d & \multicolumn{2}{c|}{Action Accuracy} \\
\hline
Approach & MPJPE [px] $\downarrow$ & Quality $\uparrow$ & top-1 $\uparrow$ & top-3 $\uparrow$ & MPJPE [px] $\downarrow$ & Quality $\uparrow$ & top-1 $\uparrow$ & top-3 $\uparrow$\\
\hline\hline
No Objects & 61 & 0.52 & 28\% & 51\% & 42 & 0.30 & 29\% & 50\% \\
No Noise & 55 & 0.49 & 29\% & 50\% & 48 & 0.29 & \textbf{30}\% & \textbf{51\%} \\
\hline\hline
\textbf{Ours} & \textbf{50} & \textbf{0.55} & \textbf{29\%} & \textbf{51\%} & \textbf{40} & \textbf{0.31} & 29\% & 50\% \\
\hline
\end{tabular}
}
\caption{Ablations studies with no object input and no noise input.}
\vspace{-2em}
\label{tab:nonoise-noobjects}
\end{center}
\end{table}

\subsection{Statistical Action Baselines}
We additionally evaluate ``Zero Velocity'' and ``Train Average'' for action labels, analogous to forecasted poses, i.e. repeating the last action label and repeating the most frequent train action label, in Tab.~\ref{tab:statistical-action-baselines}. These baselines perform particularly poorly since actions are rarely repeated or fixed for entire sequences.

\begin{table}[h]
\begin{center}
\resizebox{\columnwidth}{!}{
\begin{tabular}{|l||c|c||c|c|}
\hline
 & \multicolumn{2}{c||}{MPII Cooking II} & \multicolumn{2}{c|}{IKEA ASM} \\
\hline
Approach & top-1 $\uparrow$ & top-3 $\uparrow$ & top-1 $\uparrow$ & top-3 $\uparrow$\\
\hline\hline
Repeat Last Input & 9\% & 43\% & 8\% & 35\% \\
Most Common in Train & 6\% & 10\% & 7\% & 26\% \\
\hline\hline
\textbf{Ours} & \textbf{29\%} & \textbf{51\%} & \textbf{29\%} & \textbf{50\%}  \\
\hline
\end{tabular}
}
\caption{Statistical action baselines: (1) Repeat the last input action label (2) Using the most common action label of the train set.}
\label{tab:statistical-action-baselines}
\end{center}
\vspace{-2em}
\end{table}

\begin{table}
\begin{center}
\resizebox{\columnwidth}{!}{%
\begin{tabular}{|c|c||c|c||c|c|}
\hline
 & 2D & 3D & \multicolumn{2}{c|}{Action Accuracy} \\
\hline
Poses & MPJPE [px] $\downarrow$ & Quality $\uparrow$ & top-1 $\uparrow$ & top-3 $\uparrow$\\
\hline\hline
Uncoupled & 64 & 0.30 & 28\% & 48\% \\
\hline
Middle & 47 & 0.35 & 28\% & 47\% \\
Random & 49 & 0.24 & 28\% & 49\% \\
\textbf{Characteristic} & \textbf{41} & \textbf{0.35} & \textbf{29\%} & \textbf{50\%} \\
\hline
\end{tabular}
}
\caption{Ablation on pose forecasting, on the IKEA-ASM~\cite{ben2021ikea} dataset. We consider predicting poses following state-of-the-art pose forecasting in a decoupled fashion from actions (uncoupled), as well as poses coupled to actions in various fashions: middle (the middle pose of an action range), random (a random pose of the action), and our characteristic pose prediction, which benefits action prediction the most.}
\vspace{-2em}
\label{tab:ablations-charposes-ikea}
\end{center}
\end{table}

\section{Baseline Evaluation Details}
\label{sec:baseline-evaluation}
\subsection{State-of-the-Art Pose Forecasting}
We evaluate the performance of our baselines using the same input data that is available to our method. Pose forecasting baselines DLow~\cite{yuan2020dlow}, GSPS~\cite{mao2021gsps}, and STARS~\cite{DBLP:conf/eccv/XuWG22} are trained and evaluated on sequences of our manually annotated characteristic poses. Since there is no ground-truth 3D pose data available, we first use RepNet~\cite{wandt2019repnet}, a state-of-the-art 3D pose estimation method, to retrieve 3D skeletons from our 2D characteristic poses. We train this method from scratch using the same database of valid 3D poses that is available to our method, allowing for a fair comparison.

\subsection{State-of-the-Art Action Label Forecasting}
We train action baselines AVT~\cite{girdhar2021anticipative} and FUTR~\cite{DBLP:conf/cvpr/GongLKHC22} using sequences of our characteristic pose frames together with the corresponding action labels as input. For AVT, we use their default parameters used by the original authors for their ablation on third-person dataset 50Salads/Breakfast, inputting our RGB frames instead.
For a fair comparison, we also supply the action and object history for each step by encoding both label sequences with a small encoder (a single linear layer) each and fuse these features with the image features generated by the AVT encoder.
For FUTR, we first generate I3D features \cite{DBLP:conf/cvpr/CarreiraZ17} from our RGB frames and concatenate them with action and object history after encoding these in the same way as for AVT.

We then train two variants of both methods: One with the raw RGB frames, action history, and object history as input (``AVT RGB'' and ``FUTR RGB'' in the main results figure), and one with additional 2D skeleton input (skeletons rendered on top of the RGB frames) from the skeletons that we extract with OpenPose~\cite{cao19OpenPose} (``AVT RGB+Skeleton'' and ``FUTR RGB+Skeleton'').

\subsection{Supervised 3D Pose Lifting}
For better comparability, we used weakly supervised approach \cite{wandt2019repnet} for pose lifting. This is important, since there is no ground-truth coupling between 2D and corresponding 3D action poses in our setting. Nevertheless, we compare to baselines \cite{yuan2020dlow,mao2021gsps,DBLP:conf/eccv/XuWG22} in Tab.~\ref{tab:spin} with poses lifted using fully supervised pre-trained SPIN \cite{kolotouros2019spin}; our approach outperforms even these improved baselines in terms of 2D MPJPE.

\begin{table}[h]
\begin{center}
\resizebox{\columnwidth}{!}{%
\begin{tabular}{|l||c|c||c|c|}
\hline
 & \multicolumn{2}{c||}{MPII Cooking II} & \multicolumn{2}{c|}{IKEA ASM} \\
\hline
 & 2d & 3d & 2d & 3d \\
\hline
Approach & MPJPE [px] $\downarrow$ & Quality $\uparrow$ & MPJPE [px] $\downarrow$ & Quality $\uparrow$ \\
\hline\hline
SPIN \cite{kolotouros2019spin} + DLow \cite{yuan2020dlow} & 81 & \textbf{0.89} & 43 & \textbf{0.43} \\
SPIN \cite{kolotouros2019spin} + GSPS \cite{mao2021gsps} & 74 & 0.66 & 45 & 0.29 \\
SPIN \cite{kolotouros2019spin} + STARS \cite{DBLP:conf/eccv/XuWG22} & 66 & 0.80 & 41 & 0.40 \\
\hline\hline
\textbf{Ours} & \textbf{50} & 0.55 & \textbf{40} & 0.31 \\
\hline
\end{tabular}}
\caption{Comparison to pose baselines using fully-supervised pre-trained 3D pose estimation method SPIN~\cite{kolotouros2019spin}. In our main experiments, we instead compare to weakly supervised baseline RepNet~\cite{wandt2019repnet} for a fair comparison.}
\label{tab:spin}
\end{center}
\vspace{-2em}
\end{table}

\section{3D Quality Metric Details}
\label{sec:quality-metric}
For our pose quality metric, we use a 3-layer MLP binary classifier of 3D poses.
Training poses are randomly sampled from ground-truth (real) and predicted (fake) collected during the training process of our method and all baselines, producing a total of $100$k real and fake poses each. Fake poses exhibit a range of small to large unrealistic deformations, depending on when they were sampled, ranging from random joint placements to widely inconsistent bone lengths to unnatural joint angles to only minor inconsistencies in the bone lengths. The classifier is trained once and then used to evaluate all methods, to ensure a fair comparison.

As an additional intuitive metric we show the mean absolute bone length difference of right and left body in 3D in Tab.~\ref{tab:symmetry-quality}. We observe that this metric correlates with our classifier-based quality.

\begin{table}[h]
\vspace{-1em}
\begin{center}
\resizebox{\columnwidth}{!}{%
\begin{tabular}{|c||c|c||c|c|}
\hline
 & \multicolumn{2}{c||}{MPII Cooking II} & \multicolumn{2}{c|}{IKEA ASM} \\
\hline
Approach & Symm. [mm] $\downarrow$ & Quality $\uparrow$ & Symm. [mm] $\downarrow$ & Quality $\uparrow$ \\
\hline\hline
RepNet \cite{wandt2019repnet} + DLow \cite{yuan2020dlow} & \textbf{13} & \textbf{0.72} & 45 & 0.31 \\
RepNet \cite{wandt2019repnet} + GSPS \cite{mao2021gsps} & 18 & 0.66 & 56 & 0.15 \\
RepNet \cite{wandt2019repnet} + STARS \cite{DBLP:conf/eccv/XuWG22} & 16 & 0.62 & 46 & 0.27 \\
\hline\hline
No 3D Adversarial Loss & 75 & 0.10 & 66 & 0.05 \\
2D Projection Loss Only & 57 & 0.21 & 61 & 0.09 \\
No Action Loss & 22 & 0.53 & 39 & 0.29 \\
\hline\hline
\textbf{Ours} & 22 & 0.55 & \textbf{39} & \textbf{0.31} \\
\hline
\end{tabular}
}
\caption{Additional quality metric and its correlation to our classifier-based metric: Absolute bone length difference between right and left body, compared to pose baselines and ablations.}
\label{tab:symmetry-quality}
\vspace{-2em}
\end{center}
\end{table}

\section{Multi-Actor Interaction Scenario}
\label{sec:multi-actor}
In addition to our experiments with single human actors in the main paper, we show here that our approach is able to generalize to multi-actor scenarios, with minor modifications. We show this in Tab.~\ref{tab:multi-actor} with additional dataset TICaM \cite{DBLP:conf/bmvc/KatroliaEFMRS21} where driver and passenger are interacting in an in-car driving scenario (actions include ``talking'', various handoffs). Our modifications are: \textbf{(1)} Additional encoder and decoder for the second person \textbf{(2)} Interaction pooling introduced in Social GAN \cite{DBLP:conf/cvpr/GuptaJFSA18}. Our modified method outperforms simple combinations of previous works, with and without interaction modelling, demonstrating the wide applicability of our method.

\begin{table}[h]
\begin{center}
\begin{subfigure}{\columnwidth}
\resizebox{\columnwidth}{!}{%
\begin{tabular}{|l||c|c||c|c|}
\hline
 & 2d & 3d & \multicolumn{2}{c|}{Action Accuracy} \\
\hline
Approach & MPJPE [px] $\downarrow$ & Quality $\uparrow$ & top-1 $\uparrow$ & top-3 $\uparrow$ \\
\hline\hline
FUTR RGB + Skeleton & - & - & 38\% & 64\% \\
\hline
RepNet + STARS & 89 & 0.34 & - & - \\
\hline\hline
\textbf{Ours (No Interactions)} & 68 & 0.40 & 40\% & 67\% \\
\textbf{Ours (Interaction Modeling)} & \textbf{58} & \textbf{0.41} & \textbf{48\%} & \textbf{73\%} \\
\hline
\end{tabular}}
\end{subfigure}
\hfill
\begin{subfigure}{0.5\columnwidth}
\includegraphics[width=125px]{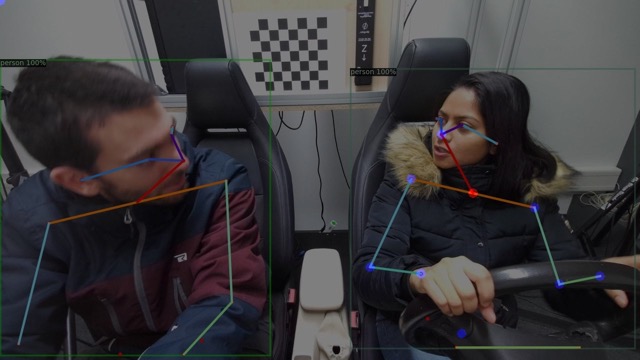}
\caption*{Setting}
\end{subfigure}
\caption{Our approach can also be applied to multi-actor scenarios: We demonstrate improved performance on suitable dataset TICaM \cite{DBLP:conf/bmvc/KatroliaEFMRS21}, with and without explicit interaction modeling.}
\vspace{-2em}
\label{tab:multi-actor}
\end{center}
\end{table}

\section{2D Input Pose Quality}
\label{sec:2d-detectors}
In Fig.~\ref{tab:pose-input-robustness}, we replace OpenPose with AlphaPose \cite{DBLP:journals/pami/FangLTXZXLL23} and Detectron2 \cite{wu2019detectron2}, both only slightly changing the final results, indicating that our method does not depend on a specific 2D pose detector. We also experiment with added random noise to OpenPose: our method remains relatively robust. The coupled changes in pose and action accuracy further demonstrate the effectiveness of our joint feature learning.

\begin{table}[h]
\begin{center}
\resizebox{\columnwidth}{!}{%
\begin{tabular}{|l||c|c||c|c|}
\hline
MPII Cooking II & 2d & 3d & \multicolumn{2}{c|}{Action Accuracy} \\
\hline
Approach & MPJPE [px] $\downarrow$ & Quality $\uparrow$ & top-1 $\uparrow$ & top-3 $\uparrow$ \\
\hline\hline
OpenPose + max. 20px noise & 59 & 0.45 & 26\% & 47\% \\
OpenPose + max. 10px noise & 57 & 0.47 & 26\% & 46\% \\
\hline
Ours (using Detectron2) & 47 & \textbf{0.54} & 28\% & 55\% \\
Ours (using AlphaPose) & \textbf{46} & 0.57 & 28\% & \textbf{56\%} \\
\hline\hline
Ours (using OpenPose) & 50 & 0.55 & \textbf{29\%} & 51\% \\
\hline
\end{tabular}}
\caption{Robustness of our method to different 2D pose detectors Detectron2 \cite{wu2019detectron2} and AlphaPose \cite{DBLP:journals/pami/FangLTXZXLL23} as well as randomly added 2D noise. This only slightly affects our pose and action accuracy, further demonstrating the effectiveness of our joint feature learning.
}
\label{tab:pose-input-robustness}
\end{center}
\vspace{-1em}
\end{table}

\section{Architecture Details}
\label{sec:architecture}
\begin{figure*}
    \centering
    \includegraphics[width=\linewidth]{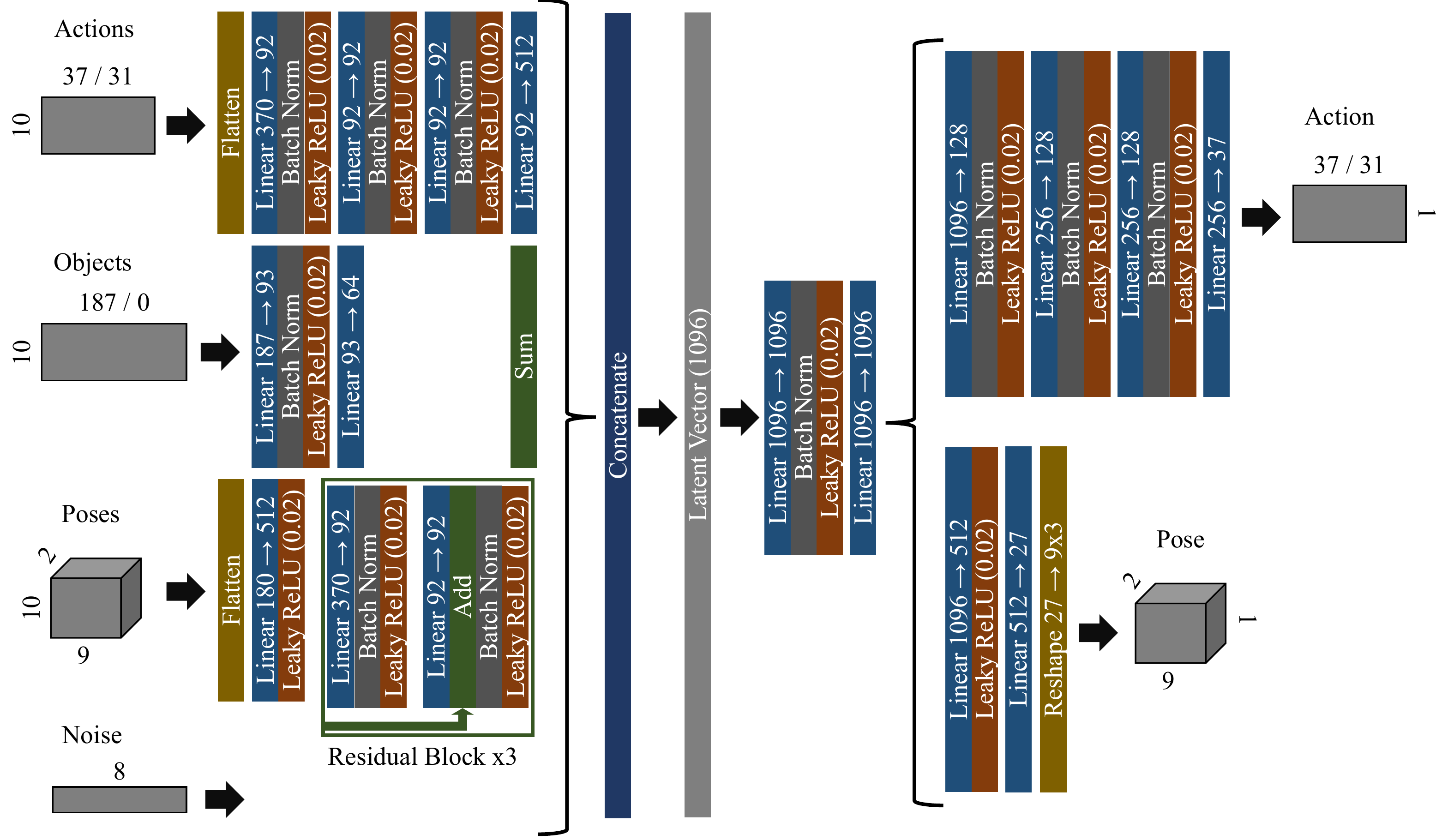}
    \caption{Network architecture specification.}
    \label{fig:architecture}
\end{figure*}

\noindent
\textbf{Generator Network}
Fig.~\ref{fig:architecture} shows our generator architecture in detail with input and output dimensions for linear layers, and the slope for leaky ReLU layers.

\smallskip
\noindent
\textbf{Critic Network}
Our adversarial critic network processes generator outputs with 4 linear layers and 3 kinematic chain layers which are designed to encourage correct bone lengths (as shown in~\cite{wandt2019repnet}), in parallel. 2 linear layers then combine both outputs and produce the final critic score.

\section{Data Details}
\label{sec:data-details}
\subsection{Camera Parameters}
While intrinsic camera parameters are often available in captured image data, the camera parameters for captured video were not available from the MPII Cooking 2~\cite{rohrbach15ijcv} dataset to use for pose projection.
We thus optimized for intrinsic camera parameters from the video sequence data in correspondence with the 3D scene reconstruction of the empty kitchen environment, as given by \cite{susanto20123d}.
For IKEA-ASM~\cite{ben2021ikea}, we use the provided intrinsic camera parameters directly.
Note that camera parameters are only required to be fixed within a sequence (i.e. no moving camera) but can change between sequences.

\subsection{3D Pose Database Alignment}
We use popular 3D pose datasets Human3.6m~\cite{DBLP:journals/pami/IonescuPOS14}, AMASS~\cite{AMASS:ICCV:2019}, and GRAB~\cite{GRAB:2020} for our database of uncorrelated valid 3D poses. All poses are pre-processed to follow the OpenGL coordinate system and aligned with respect to the neck joint.

\subsection{Pose Joint Layout}
We use the 9 upper-body joints of the native OpenPose \cite{cao19OpenPose} joint layout for skeletons in 2D, and adapt skeletons in our 3D database to use the same format. Tab.~\ref{tab:pose-layout} shows the correspondence between our joint layout, OpenPose \cite{cao19OpenPose}, Human3.6m \cite{DBLP:journals/pami/IonescuPOS14}, and SMPL-X \cite{SMPL-X:2019}. 3D datasets AMASS \cite{AMASS:ICCV:2019} and GRAB \cite{GRAB:2020} provide human bodies in SMPL-X format; we first extract their skeleton joints using their publicly available code and then convert it into our layout using the correspondences in Tab.~\ref{tab:pose-layout}.

\begin{table}
\centering
\resizebox{\columnwidth}{!}{%
\begin{tabular}{|c|c||c|c|c|c|c|c|}
\hline
    \multicolumn{2}{|c||}{Ours} & \multicolumn{2}{|c|}{OpenPose} & \multicolumn{2}{|c|}{Human3.6m} & \multicolumn{2}{|c|}{SMPL-X} \\\hline
     Idx & Name & Idx & Name & Idx & Name & Idx & Name \\\hline
     0 & head & 0 & nose & 15 & head & 15 & head \\
     1 & neck & 1 & neck & 13 & thorax & 12 & neck \\
     2 & right shoulder & 2 & right shoulder & 25 & right shoulder & 17 & right shoulder \\
     3 & right elbow & 3 & right elbow & 26 & right elbow & 19 & right elbow \\
     4 & right hand & 4 & right hand & 27 & right wrist & 42 & right index 3\\
     5 & left shoulder & 5 & left shoulder & 17 & left shoulder & 16 & left shoulder \\
     6 & left elbow & 6 & left elbow & 18 & left elbow & 18 & left elbow \\
     7 & left hand & 7 & left wrist & 19 & left wrist & 27 & left index 3 \\
     8 & hip & 8 & mid-hip & 0 & hip & 0 & pelvis \\
     \hline
\end{tabular}
}
\caption{Human skeleton joint layout used in our experiments, for both 2D and 3D skeletons.}
\label{tab:pose-layout}
\end{table}

\subsection{MPII Cooking 2 Details}
We use action labels as annotated in the 2D cooking action dataset MPII Cooking 2~\cite{rohrbach15ijcv}. These annotations provide action labels (87 classes) for frame ranges in each sequence as well as the involved objects (187 classes).
We first cluster similar actions together, yielding a total of 37 action clusters, which we then use as action classes in our experiments.

In addition, since our goal is to forecast upper-body actions with objects in the foreground, we remove instances of poses and corresponding actions that occur in the background - e.g., when taking out objects from the cupboard, or from the fridge.

In total, there are 272 cooking action sequences; we create a random train/val/test split along sequences with a ratio of 70\% / 15\% / 15\%, yielding 190, 40, 40 sequences for each set.

\subsection{IKEA-ASM Details}
We use action labels as annotated in the IKEA furniture assembly dataset IKEA-ASM~\cite{ben2021ikea}. These annotations provide action labels (31 classes) for frame ranges in each sequence; we use them without explicit object information since each action already encodes its associated object.

In total, there are 370 furniture assembly action sequences; we create a random train/val/test split along sequences with a ratio of 70\% / 15\% / 15\%, yielding 227, 48, 48 sequences for each set.

\end{appendices}

\end{document}